\documentclass{article}
\usepackage{natbib}

\usepackage{PRIMEarxiv}

\usepackage[utf8]{inputenc} % allow utf-8 input
\usepackage[T1]{fontenc}    % use 8-bit T1 fonts
\usepackage{hyperref}       % hyperlinks
\usepackage{url}            % simple URL typesetting
\usepackage{booktabs}       % professional-quality tables
\usepackage{amsfonts}       % blackboard math symbols
\usepackage{nicefrac}       % compact symbols for 1/2, etc.
\usepackage{microtype}      % microtypography
\usepackage{pifont} % for \XSolidBrush and checkmark symbols
\usepackage{lipsum}
\usepackage{natbib}
\usepackage{epstopdf}
\usepackage{fancyhdr}       % header
\usepackage{graphicx}       % graphics
\usepackage{amssymb}
\usepackage{multirow}
\usepackage{bbding}

\usepackage{threeparttable}
\graphicspath{{media/}}     % organize your images and other figures under media/ folder

\title{Feature-Enhanced TResNet for Fine-Grained Food Image Classification
\thanks{\textit{\underline{Citation}}: 
\textbf{Lulu Liu, Zhiyong Xiao. Feature-Enhanced TResNet for Fine-Grained Food Image Classification }} 
}

\author{
  Lulu Liu\\\\
  School of Artificial Intelligence and Computer Science 
  Jiangnan University \\
  Wuxi, 214122, China\\
   \And
  Zhiyong Xiao* \\\\
  School of Artificial Intelligence and Computer Science 
  Jiangnan University \\
  Wuxi, 214122, China \\
  \texttt{zhiyong.xiao@jiangnan.edu.cn} \\
}

\begin{document}
\maketitle

\begin{abstract}
Food is not only essential to human health but also serves as a medium for cultural identity and emotional connection. In the context of precision nutrition, accurately identifying and classifying food images is critical for dietary monitoring, nutrient estimation, and personalized health management. However, fine-grained food classification remains challenging due to the subtle visual differences among similar dishes. To address this, we propose Feature-Enhanced TResNet (FE-TResNet), a novel deep learning model designed to improve the accuracy of food image recognition in fine-grained scenarios. Built on the TResNet architecture, FE-TResNet integrates a Style-based Recalibration Module (StyleRM) and Deep Channel-wise Attention (DCA) to enhance feature extraction and emphasize subtle distinctions between food items. Evaluated on two benchmark Chinese food datasets—ChineseFoodNet and CNFOOD-241—FE-TResNet achieved high classification accuracies of 81.37\% and 80.29\%, respectively. These results demonstrate its effectiveness and highlight its potential as a key enabler for intelligent dietary assessment and personalized recommendations in precision nutrition systems.
\end{abstract}

% keywords can be removed

\section{Introduction}
The rapid advancement of computer vision technology has led to the broad application of image classification techniques across a spectrum of fields, including biometric identification, product categorization, and the stringent monitoring of food safety. Fine-grained image classification, a particularly complex and specialized task, has emerged as a pivotal area with significant potential in these critical applications \citep{r2,r3}. Unlike conventional image classification, fine-grained classification presents significant challenges due to the high degree of similarity between categories, including the nuanced differences in the appearance of various fruit types and the pronounced distinctions among individuals within the same category, as evidenced by the significant variations in color, size, and shape among different apple cultivars, adding to the complexity of the classification endeavor \cite{r0}. Food image classification highlights the inherent challenges in fine-grained classification \cite{XIAO2025}, as complexity is not limited to the food items themselves; the context provided by the background and the perspective determined by the camera angle also exert a profound influence on classification outcomes. Traditional research in food image classification largely relies on manual feature extraction methods \cite{r4,r5,r11,Xiao2014}, that often fail to capture the full spectrum of food characteristics, thereby limiting classification accuracy. In contrast, deep learning techniques, leveraging the hierarchical structure of neural networks, can automatically extract intricate features from food images, significantly enhancing classification precision \cite{r6,r7,r8}.

\begin{figure*}[!ht]
	\centering	
	\includegraphics[width=16cm,height=8cm]{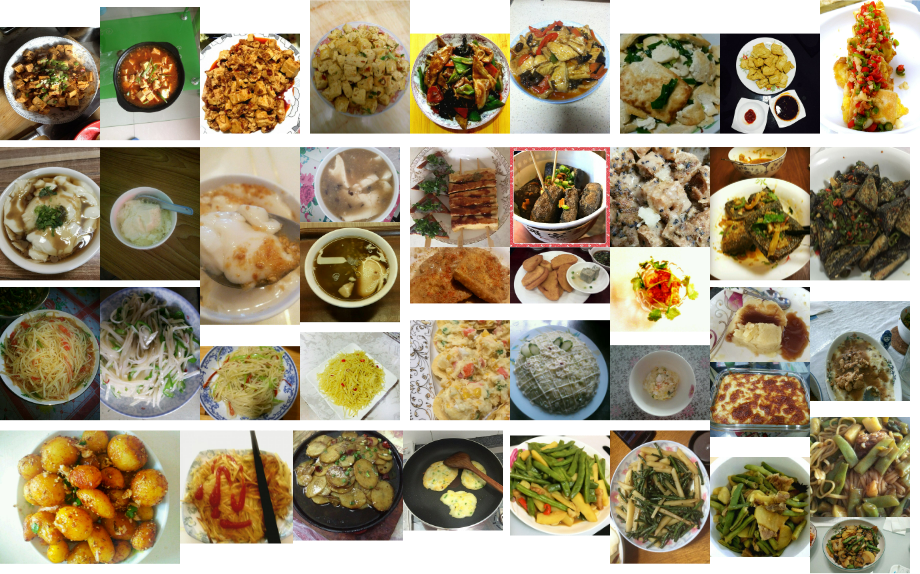}
	\caption{These are example images from food dataset. The above two lines are pictures of tofu. Due to different cooking and ingredients, different tofu foods are presented, from left to right, from top to bottom, in order: Mapo tofu, home-style tofu, fried tofu, tofu flower and stinky tofu; The bottom two lines are potato pictures, which are successively hot and sour shredded potatoes, mashed potatoes, fried potatoes and potato braised beans. Besides, they are placed in different utensils. Different shooting environments and other factors lead to changes in the visual appearance of Chinese food images, which brings challenges to visual food classification.}
	\label{p4}
\end{figure*}

Since the advent of deep learning, it has received widespread attention from researchers. Several robust models, such as ResNet \cite{r12,r13,r14}, Vision Transformer (ViT) \cite{r51}, Inception \cite{r35}, EfficientNet \cite{r36}, and MobileNet \cite{r37,r38}, have been developed and widely applied to tasks such as image segmentation \cite{r24,Qian2020, BSPC2021, CMPB2021, FoodSeg2025,Liu20223}, classification \cite{XIAO2025pestclassification}, and recognition \cite{XIAO2025,Wang2025}. Zhiyong Xiao et al. introduced an innovative semi-supervised learning optimization strategy based on the teacher-student paradigm that integrates the strengths of Convolutional Neural Networks (CNNs) and Transformers \cite{r46}. This integration significantly enhanced the efficiency of utilizing large volumes of unlabeled medical images and improved the efficacy of model segmentation outcomes. Chao Ji et al. proposed an MLP-based model employing matrix decomposition and rolling tensor technology for skin lesion segmentation, replacing the self-attention mechanism of Transformers \cite{r47}. The research not only resulted in superior model performance with a reduced parameter count but also demonstrated the innovative application of traditional methods can lead to groundbreaking results in the field of deep learning. Moreover, the adaptability of certain models to image segmentation tasks has been elegantly demonstrated through innovative modifications. This approach merged the capabilities of CNNs and ViTs to craft the Light3DHS model, designed for the segmentation of 3D hippocampal structures \cite{r50}. The Light3DHS model not only refined segmentation accuracy but also made a substantial impact on brain disease research, fostering deeper clinical investigations.

Recently, in the realm of food image classification and recognition, Xinle Gao et al. successfully tackled the intricate task of classifying food images with similar morphologies using sophisticated data and feature enhancement strategies, with the Vision Transformer (AlsmViT) \cite{r48}. Also in 2024, Zhiyong Xiao et al. introduced a deep convolutional module that, after being integrated into the comprehensive feature representation derived from Swin Transformer, notably enhanced the depth and intricacy of both local and global feature representations \cite{r49}. Translating the hierarchical parallelism and evolutionary optimization philosophy of hybrid parallel genetic algorithm to food classification can transcend the limitations of conventional CNNs, particularly excelling in scenarios with class-imbalanced and fine-grained food recognition \cite{HPGA}. A lightweight fine-grained food recognition model integrates an FIP-Attention module for modeling complex ingredient-dish relationships and an FCR-Classifier for refining texture-color features, achieving state-of-the-art performance on multiple benchmark datasets and enabling practical deployment in a mobile dietary monitoring application \cite{FGFoodNet}.

Building on these seminal models, the TResNet model has been introduced, achieving a reduction in both the number of parameters and computational complexity (FLOPs). This model sustains high efficiency in GPU-based training and inference processes and offers increased throughput, significantly boosting the accuracy of neural networks. TResNet enhances ResNet50's high throughput and residual architecture and further incorporates the Squeeze-and-Excitation (SE) attention mechanism \cite{r9}. This mechanism enables adaptive learning of channel-wise features within the neural network while retaining the original features, thereby further enhancing TResNet's performance. TResNet has become a popular method for computer vision tasks \cite{r23,r25}, addressing multi-label classification and fine-grained partitioning \cite{r28,r29}. In a study by Zelin Xu et al. on the diagnosis of Alzheimer's Disease, TResNet was employed as the backbone network, with SK adjustments replacing SE to dynamically adjust the size of the Convolutional kernel for capturing features from various receptive fields, achieving an 86.9\% accuracy rate in AD diagnosis \cite{r22}. Later, Changhyun Kim et al. combined TResNet and Feature Pyramid Network (FPN) to process multi-label classification tasks in medical images, demonstrating robust performance across different lesion sizes \cite{r26}. In recent work, Ji-Hyeon Lee et al. utilized a Gaussian filter to remove noise from imaging results in brain tumor classification within magnetic resonance imaging and applied the Patterned-GridMask method in TResNet, achieving a classification accuracy of 97.74\% \cite{r28}. The TResNet model is not only widely useful in medicine but also well represented in the category of fine granularity. Also in 2023, Dichao Liu et al. explored fine-grained visual classification on FGVC-Aircraft, Stanford Cars, and Food-11, achieving good accuracy \cite{r31}.

In fact, not every part of an image holds the same significance for classification tasks. Critical is pinpointing and concentrating on regions that are intimately linked to the classification objectives, as well as capturing the interdependencies among pixels. Within Convolutional Neural Networks (CNNs), the Convolutional operation primarily focuses on areas within a confined receptive field, neglecting the crucial connections that distant pixels may share. Thus, capturing long-range dependencies is vital within the architecture of neural networks. In the realm of fine-grained visual categorization, attention mechanisms have emerged as a pivotal subject of exploration, supporting models in surmounting the adversities stemming from intrinsic similarities between classes and variations within a class, and concentrating on the critical details within localized areas \cite{r30}. This study introduces a novel image classification approach termed FE-TResNet, predicated on the TResNet backbone network to enhance feature extraction and multi-scale feature fusion more effectively. The FE-TResNet network comprises two core components: a feature extractor and a feature classifier. The feature extractor includes three distinct parts: the backbone network feature extractor, StyleRM, and DCA. Enhancing the features derived from the backbone network feature extractor with stylized weights allows us to determine feature importance through varying stylized weight values, thereby enhancing fine-grained control over feature stylization. Furthermore, multi-scale feature fusion is achieved by combining features from different channels, preserving spatial dimensions while integrating features from various directions. This approach not only improves the model's ability to discern global and local features but also enriches the dimensionality of feature representation. For the feature classifier, a generic fine-grained food classifier is utilized to process features that have been processed by the feature extractor. Experiments were conducted on fine-grained food image classification using the ChineseFoodNet and CNFOOD-241 datasets with the FE-TResNet model. The experimental results showed that the FE-TResNet model achieved classification accuracies of 81.37\% and 80.29\% on these two datasets, respectively. These results significantly outperform existing techniques, thereby validating the model's exceptional performance and potential in image classification tasks.

\textbf{To summarize, our key contributions are:}
\begin{itemize}
\item This paper proposes a customized FE-TResNet network to achieve fine-grained image classification.
\item StyleRM resolves several key issues, including low utilization of stylization feature extraction parameters, limited capture of complex stylization features, and loss of spatial information. This adaptation facilitates the model's processing of the texture and structure of the input data.
\item DCA enables multi-scale feature fusion by skillfully combining information from different channels. This strategy not only captures objects and details that may be overlooked during the convolution process but also effectively meets the demand for a large number of pixel points in the processing of high-resolution images. Consequently, DCA significantly enhances the flexibility of the network model, enabling more efficient processing and analysis of image data, and improved performance in image recognition and classification tasks.
\end{itemize}

This paper is explained by the following structure. Section II introduces the structure of the FE-TResNet model. In section III, dataset is used for image classification experiments. Section IV overviews the results and analysis of ablation experiments are introduced in detail. lastly, Section V concludes a summary of the full text and the prospect of future work.

\section{Related work}
\label{}
\subsection{TResNet}
\begin{figure*}[!ht]
	\centering	
	\includegraphics[width=16cm,height=14cm]{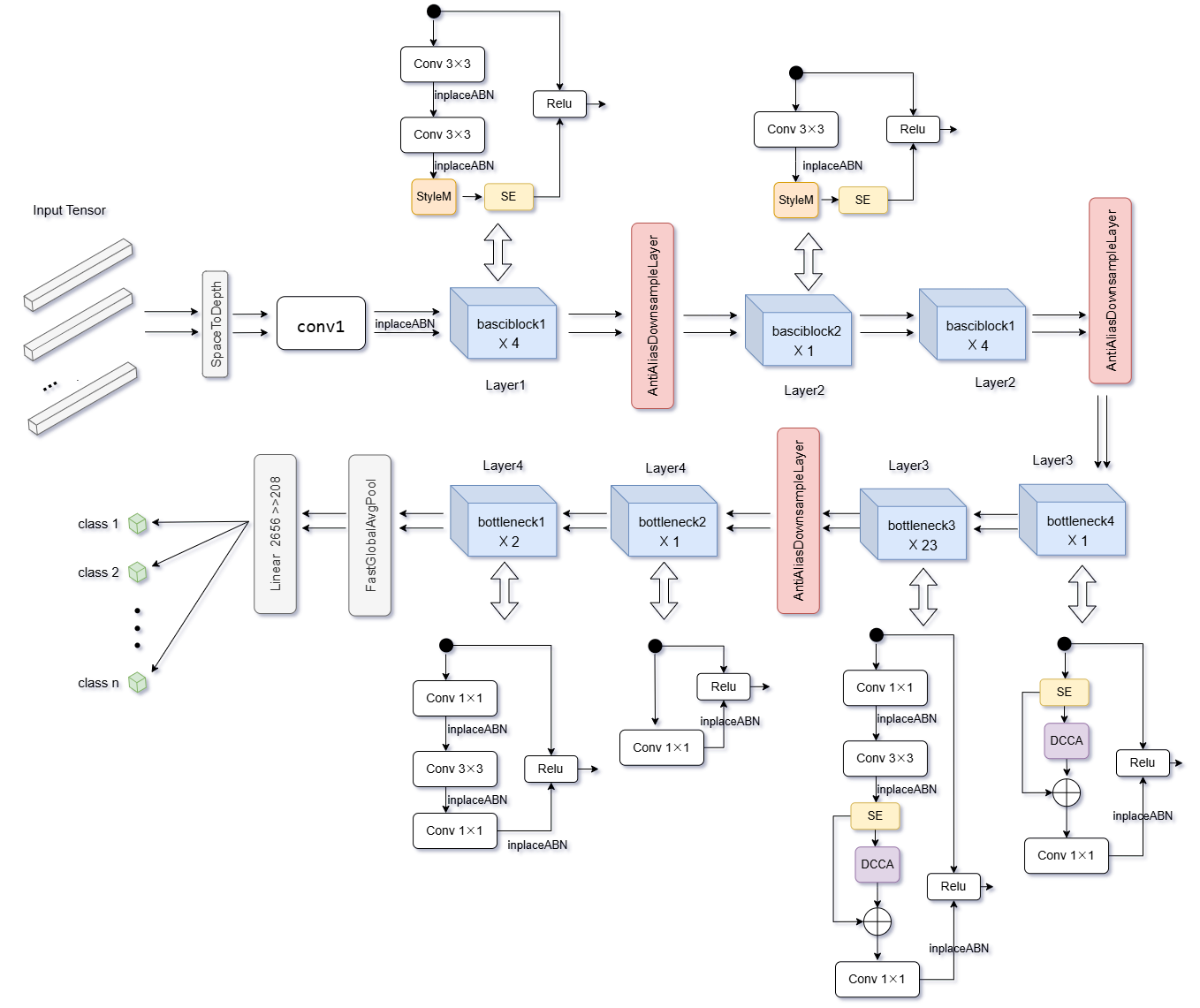}
	\caption{Overall structure of the model. Detailed design of Basicblock and Bottleneck series blocks. The two on the top are basicblock blocks containing the StyleRM and SE attention modules; The four on the footer are bottleneck blocks, and the bottleneck block on the right includes SE and DCA.}
	\label{p5}
\end{figure*}

Since AlexNet's inception, neural network architectures have trended toward increased depth. The common belief was that deepening the network would proportionally enhance its feature extraction capabilities and, consequently, boost model accuracy. Yet, experimental findings have indicated that there is a threshold beyond which model precision plateaus. This problem is further compounded by escalating training and testing errors, along with the phenomena of gradient vanishing and network degradation. This insight led to a shift in focus toward widening the network's breadth. The ResNet model was developed in response to these challenges, not only expanding the network's width but also pioneering the introduction of residual connections. These connections effectively avoid the pitfalls of gradient disappearance and network degradation inherent in deep neural networks, paving the way for the creation of even more profound architectures. ResNet rapidly became the foundational model for a variety of computer vision tasks. With the evolution of model variants, there was a collective push to refine residual models further, with the dual goals of lessening the GPU throughput requirements and accelerating computational velocity. The TresNet architecture represents a monumental leap from the ResNet50 model, boasting a 3\% improvement in recognition accuracy on the ImageNet dataset. The TresNet architecture introduces three diverse architectural variations, distinguished by their depth and channel count \citep{r1}, with the most extensive parameter set totaling 77.1M. To further refine this framework, this study optimized the Novel Block-Type Selection layer by incorporating a single convolutional layer for feature extraction post the Anti-Alias Downsampling layer. This optimization not only preserves the integrity of the original features but also reduces the parameter count by 36,022, significantly boosting the model's operational efficiency. A detailed diagram illustrating the components of the model is shown in Figure \ref{p5}.

\subsection{StyleRM}
Since the studies by Gatys et al., research has overwhelmingly shown that Convolutional Neural Networks (CNNs) are adept at processing not just the morphological aspects of diverse images but also the intrinsic textural attributes, commonly referred to as styles \citep{r15,r17,r16}. Recent studies have revealed the intimate nexus between stylistic elements and the CNNs' feature representation, underscoring the pivotal role that textural characteristics play in the mechanism of feature extraction within CNNs \citep{r18,r19}. In 2019, Huang et al. proposed the Style-based Recalibration Module for extracting style features, which dynamically assesses the importance of each style and reweights the feature maps accordingly, guiding the network to focus on meaningful style features while ignoring those that are less significant \citep{r10}. The module operates through two key steps: Style Pooling and Style Integration. Style Pooling consolidates feature responses across spatial dimensions through global average pooling and global standard deviation pooling, extracting style features from each channel. Style Integration then employs fully connected layer operations to generate specific style weights based on the extracted style features, and by reintegrating the feature maps, it emphasizes or suppresses these feature information, enhancing the representational capacity of CNNs. However, adjusting the stylization of features in each channel by multiplying with a single global content feature coefficient (cfc), a method that, while simple, fails to fully leverage the network's parameter potential. Furthermore, the stylized feature extraction process relies solely on basic multiplication and batch normalization operations, eventually compressed into a 1$\times$1 feature map, which limits the model's ability to capture complex style features and may lead to the loss of spatial information. To address these limitations, this study implemented convolutional layers on the original basis to capture spatial information in feature maps. This enhancement not only allows the model to discern more complex stylized features more deeply but also retains important spatial information, thus better understanding and processing the texture and structure in the input data. Furthermore, this approach reduces the risk of overfitting and enhances the overall performance of the model.

\begin{figure*}[!ht]
	\centering	
	\includegraphics[width=12cm,height=6cm]{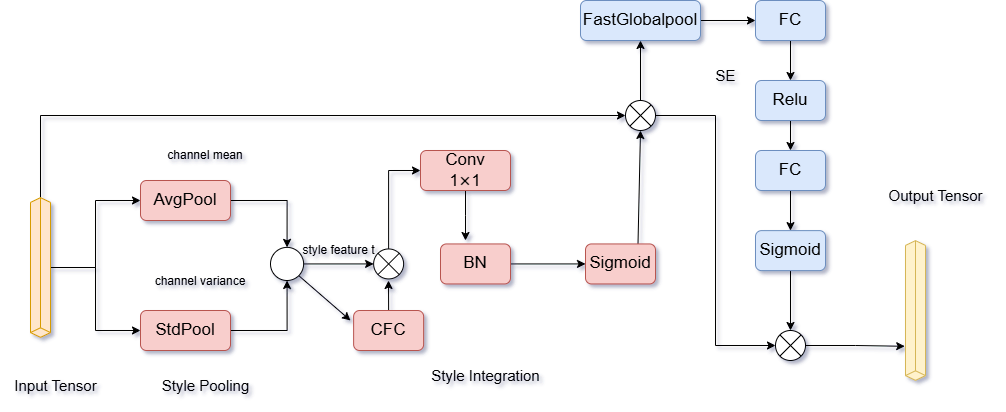}
	\caption{Detailed design diagram of the StyleRM and SE modules in series.}
	\label{p6}
\end{figure*}

\subsection{DCA}
Capturing internal interdependencies within deep neural networks is of paramount importance. In the context of audio and linguistic data, recurrent operations have been the predominant strategy for addressing these dependencies. For image data, this is typically achieved through the application of Convolutional operations in a deeply stacked and repetitive framework. However, these conventional methods are often constrained to processing dependencies within localized spatial or temporal scopes, posing inherent limitations. Essentially, there are interdependencies present between pixels that are spatially remote within an image, and these long-range dependencies are capable of encapsulating valuable contextual insights. To address this challenge, the concept of Non-Local operations was conceived \citep{r20}. This concept incorporates an efficient, simplistic, and universally applicable non-local operational mechanism that directly seizes long-range dependencies by assessing the mutual interactions between any two points in the image, irrespective of their spatial arrangement. The principle underlying Non-Local operations is that the response at any given pixel within the feature map is an aggregate of the weighted features from all other pixels, leading to considerable computational complexity of O((H$\times$W)$\times$($H\times$W)), with H and W representing the dimensions of the feature map. To enhance the Non-Local approach, the Criss-Cross operation has been developed. This operation introduces a deviation from the standard Non-Local methodology by restricting each pixel's associations to only H+W-1 points within its corresponding row and column, rather than engaging with every pixel across the map. Utilizing two sequential Criss-Cross operations serves as an effective surrogate for the Non-Local operation, facilitating the acquisition of comprehensive contextual information while significantly lowering the computational complexity to O(($H\times$W)$\times$(H+W-1)). However, this method is not sufficiently efficient when processing high-resolution images that require more pixel points, and it struggles to capture small objects and details in the images. Expanding upon CCA, this study introduce the technique of depthwise separable convolution, which decomposes standard convolution into depthwise convolution and pointwise convolution. This enables each input channel to independently apply a convolutional kernel and merge features from different channels, achieving multi-scale feature fusion. This facilitates more effective capture of directional features in the images while enhancing the perception of local features. This enhancement not only improves the model's recognition accuracy for small objects and details but also enhances the overall model performance.

\begin{figure*}[!ht]
	\centering	
	\includegraphics[width=14cm,height=10cm]{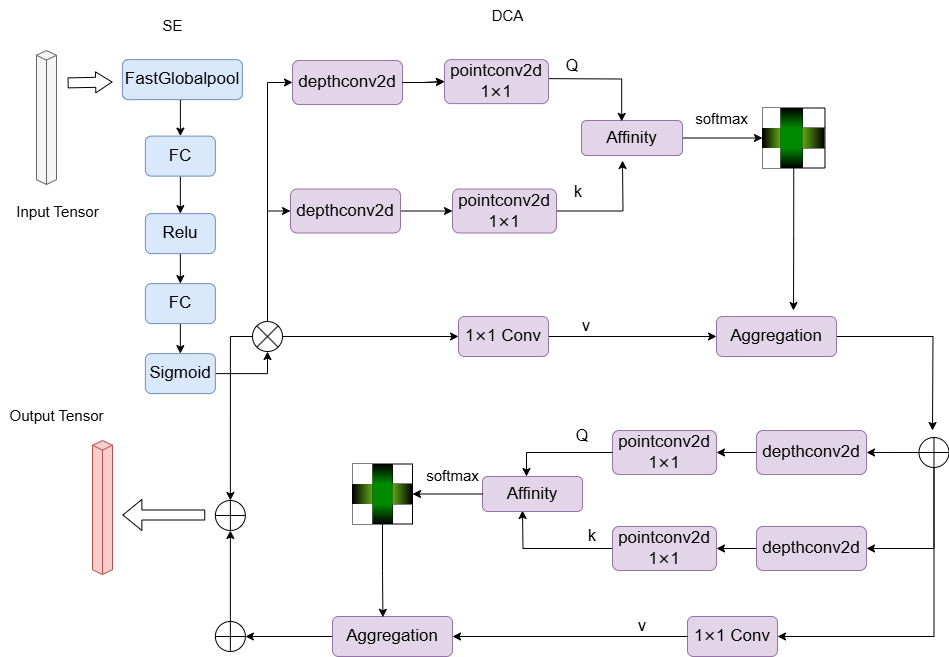}
	\caption{Detailed design diagram of the SE and DCA modules connected in series.}
	\label{p7}
\end{figure*}
\section{Materials and method}
\label{}
\subsection{Data Preparation}
This study delves into two formidable large-scale food image datasets: ChineseFoodNet \citep{r21} and CNFOOD-241 \citep{r32}. Both datasets have been generously made publicly available by the Shenzhen Institutes of Advanced Technology, Chinese Academy of Sciences, and the University of the Chinese Academy of Sciences for academic research. They have been meticulously divided into training and testing sets and can be accessed through specified links. It should be emphasized that while these datasets are intended for scholarly investigation, they are not to be employed for any commercial use without the appropriate rights and permissions. An example food image is shown in Figure \ref{p4}. The ChineseFoodNet dataset contains 185628 food images across 208 distinct categories \citep{r21}. Comprising images sourced from online recipes and menus, as well as photographs of real dishes and menus captured from everyday life \citep{r21}. The images not only demonstrate the rich culinary culture of China but also reflect the dietary habits of the Chinese people, highlighting the diversity and evolution of Chinese cuisine.

The CNFOOD-241 dataset, developed based on ChineseFoodNet, is a comprehensive dataset of Chinese food images that includes 241 categories with a total of 191811 images, all standardized to a size of 600$\times$600 pixels. The Chinese cuisine is categorized into five main groups based on the combination of meat and vegetables used in the dishes: mixed meat and vegetables, staple foods, meats, vegetarian diets, and soups. Specifically, the mixed meat and vegetables group comprises 31 types, representing 16\% of the dataset. The staple foods group contains 46 types, accounting for 22\%. The meat dishes group consists of 67 types, making up 32\% of the dataset. The vegetarian group includes 46 types, also constituting 22\%. Finally, the soup group contains 19 types, which account for 9\% \citep{r32}.

\subsection{The proposed network structure}
The feature extractor of the FE-TResNet model consists of the backbone network feature extractor, StyleRM, and DCA. Sample images are fed into the backbone network, and after convolutional processing by the basic blocks of the backbone network, the feature information is mapped to the StyleRM module. Style weights of varying magnitudes are applied to the original feature maps to capture more complex features. Subsequently, the feature maps undergo processing by the remaining SE (Squeeze-and-Excitation) blocks and activation functions of the backbone network's basic blocks to further learn deeper feature representations. As this process unfolds, the model learns richer feature representations while reducing the risk of overfitting. Thereafter, the feature information resulting from convolutional processing by the bottleneck blocks is mapped to the DCA for local feature enhancement, fusing global and local features. Ultimately, following processing by the remaining structure of the bottleneck blocks, the model completes the fusion of multi-scale features while maintaining spatial dimensions, providing greater flexibility to the model. The design of the FE-TResNet model's feature extractor more effectively manages intra-class variability and inter-class similarity in fine-grained food images, achieving more accurate category prediction. This novel methodology offers a new solution for the field of image classification, particularly in food image recognition. The FE-TResNet model structure is shown in Figure \ref{p5}.

\subsection{The fusion of StyleRM and SE}
In the realm of style transfer learning, this study employ an innovative approach to extract style information from intermediate Convolutional mappings. By leveraging the mean and standard deviation to perform dimensionality reduction on the feature maps, this study extract stylistic information from each channel. Subsequently, through a concise and efficient combination involving a Convolutional layer, batch normalization, and activation functions, this study integrate and adjust the weights for each channel to recalibrate the feature maps effectively. This process not only enhances the model's ability to capture stylistic features but also improves the outcomes of style transfer. To be specific, given input feature maps $\mathop x \in \mathop\mathbb{X}  \nolimits^{B \times C \times H \times W} $ the style features $\mathop t \in \mathop\mathbb{X}  \nolimits^{B \times C \times 2} $ are calculated by:
\begin{equation}
    \mathop \beta\nolimits_bc = = \frac{1}{HW} \sum_{h=1}^{H} \sum_{w=1}^{W}\mathop x \nolimits_bchw
\end{equation}
\begin{equation}
    \mathop \alpha \nolimits_bc = \sqrt{ \frac{1}{HW} \sum_{h=1}^{H}\sum_{w=1}^{W}{(\mathop x \nolimits_bchw-\mathop \beta\nolimits_bc)}^2} 
\end{equation}
\begin{equation}
    \mathop t\nolimits_bc=\left [ \mathop \beta\nolimits_bc, \mathop \alpha \nolimits_bc\right ] 
\end{equation}
Given the style representation $\mathop t \in \mathop\mathbb{X}  \nolimits^{B \times C \times 2} $as an input, the style integration operator performs channel-wise encoding using learnable parameters $\mathop w \in \mathop\mathbb{X}  \nolimits^{C \times 2} $, therefore, $\mathop z\nolimits_bc   = {\mathop w\nolimits_c} \times {\mathop t\nolimits_bc} $ where $\mathop z \in \mathop\mathbb{X}  \nolimits^{B \times C \times 1} $ represents the encoded style features. Then use BN and activation function operations to get the style weight, the specific implementation formula:
\begin{equation}
    \mathop \beta\nolimits_c\nolimits^{(z)}=\frac{1}{N}\sum_{n=1}^{N} \mathop z\nolimits_bc
\end{equation}
\begin{equation}
    \mathop \alpha\nolimits_c\nolimits^{(z)}=\sqrt{\frac{1}{N}\sum_{n=1}^{N}{(\mathop z\nolimits_bc -\mathop \beta\nolimits_c\nolimits^{(z)})}^2} 
\end{equation}
\begin{equation}
    \mathop z\nolimits_bc=\mathop  \gamma\nolimits_c\left ( \frac{\mathop z\nolimits_bc- \mathop \beta\nolimits_c\nolimits^{(z)}}{\mathop \alpha\nolimits_c\nolimits^{(z)}}  \right )+\mathop \beta \nolimits_c    
\end{equation}

where $\gamma, \beta \in \mathop\mathbb{X}  \nolimits^{c}$ are affine transformation parameters, and $\mathop g\nolimits_bc$ represents the channle-wise style weights. Finally, the initial input x is recalibrated with $\mathop g\nolimits_bc$ feature maps to emphasize or suppress their information. The Squeeze and Excitation block is a computing unit. $\mathop x \in \mathop\mathbb{X}  \nolimits^{B \times C \times H \times W} $ can be constructed based on  $\mathop x' \in \mathop\mathbb{X}  \nolimits^{B \times C' \times H' \times W'} $. Firstly, spatial feature compression is performed on the feature map to realize global average pooling in spatial dimension. The feature map of C$\times$1$\times$1 is obtained, and a weight vector Y is obtained through two fully connected layers and an activation function. Finally, the weight vector and the original data are multiplied x$\times$Y, and the feature map of the channel with important style features is finally obtained. The structure of the module is shown in Figure \ref{p6}

\subsection{The fusion of SE and DCA}
The squeezed and Excitation block trained feature graph is taken as the input $\mathop R \in \mathop\mathbb{X}  \nolimits^{B \times C \times H \times W} $ of the first DCA model. First, two feature graphs Q and K are obtained by two deep separable convolution1$\times$1. After affiniity and softmax, this study get an attention map A. The implementation is as follows, the spatial dimension of Q and K is $\mathop\mathbb{X}  \nolimits^{C' \times H \times W} $, where $C'< C$, at every position u in Q, this study can get a vector $\mathop Q\nolimits_u \in \mathop\mathbb{X} \nolimits^{ C'}$, meanwhile, in K, With u on the same line can get on the same column $H + W - 1$ the characteristics of the point set $\mathop K\nolimits_u \in \mathop\mathbb{X} \nolimits^{(H+W-1) \times C'}$, Affiniity formula for $\mathop D\nolimits_u = \mathop Q\nolimits_u  \times \mathop K\nolimits_u\nolimits^{\rm{T}} $, where $\mathop D \in \mathop\mathbb{X}  \nolimits^{(H+W-1) \times H \times W} $, Applying the softmax function on the D channel gives the attention map $\mathop A \in \mathop\mathbb{X} \nolimits^{(H+W-1) \times H \times W} $. The feature adaptive graph $\mathop V \in \mathop\mathbb{X}  \nolimits^{C \times H \times W} $ obtained by the third 1$\times$1 Convolution in the initial H, similarly, $\mathop V\nolimits_u$ and $\mathop V\nolimits_{u'}$($\mathop V\nolimits_{u'}$ are feature sets of $H+W-1$ points on the same row and column as u can be obtained. Finally, through the Aggregation operation, the $\mathop V\nolimits_u \times \mathop A\nolimits_u$ of each ${u'}$ on the feature graph V is added to the initial feature R, so that ${R'}$ is obtained (the sparse connection of each pixel and the global feature). If you want to achieve a dense connection like non-local, you will perform another DCA operation, taking the above ${R'}$ as R, to obtain ${R''}$ captures the long distance dependence of all pixels. Because the two DCA operations share parameters, the resulting feature map is dense and rich in context information without generating additional parameters. The module structure is shown in Figure \ref{p7}.

\subsection{Deeper architectural layers}
Deeper networks tend to perform better prior to encountering depth-related bottlenecks in model architecture. However, as the depth of networks increases, challenges such as vanishing and exploding gradients emerge, complicating the training process. Consequently, employing a diverse array of strategies and more sophisticated optimization algorithms is essential to bolster the efficiency of deep learning networks. Gao Huang et al. introduced the innovative concept of stochastic depth \citep{r42}. The approach begins training by utilizing a profoundly deep network, strategically discarding a random subset of layers in each mini-batch and circumventing them with an identity function, elegantly overcoming the limitations of residual networks in terms of depth. In 2018, Zijun Zhang et al. advanced the field with the introduction of the normalized direction-preserving Adam (ND-Adam) method \citep{r40}. The technique fine-tunes the direction and magnitude of weight updates with greater precision than Adam, effectively mitigating the generalization gap. The following year, Liyuan Liu et al., recognizing issues with adaptive learning rates, proposed warmup works as a variance reduction strategy, thereby enhancing its efficacy and robustness \citep{r43}. In recent work, Hao Shao et al. have enhanced deep classification by integrating a linear enhancement of logits within the Softmax function, fostering intra-class compactness and inter-class divergence, and thus, elevating the model's generalization in classification tasks \citep{r41}.

\section{Experiment and Result analysis}
\label{}
\subsection{Experiment preparation}
The experiments were conducted using the deep learning framework provided by the PyTorch library, executed on an NVIDIA GeForce RTX 2080Ti GPU. The datasets used were carefully split into training and validation sets. Before training on the training set, input images were randomly cropped and scaled to match the uniform pixel size required by the model. This approach ensures that the model observes different parts of the images with each iteration, effectively preventing overfitting. Moreover, this study implemented automated data augmentation via AutoAugment, diversifying the training dataset through a combination of rotations, cropping, and scaling. Regarding the validation set, after adjusting the input images to a fixed size, a region was cropped from the center using the CenterCrop transformation. These preprocessing steps facilitated the application of image data to the model training. Additional experimental parameters are detailed in Table \ref{tab:one}.

\begin{table*}[!ht]
\scriptsize
\centering
\caption{Parameters in the experimental process}
\label{tab:one}
\setlength{\tabcolsep}{2mm}{
\begin{threeparttable} 
\begin{tabular}{lllllllllll}
\hline
Task     & Model   & Datasets       & Input & Epochs & Batch size & Optimizer & LR     & LR decay & weight decay & Warmup epochs \\ \hline
Pretrain & TResNet & ImageNet-21K   & 224   & 300    & 24         & SGD       & 1.E-01 & Cosine   & 0.01         & 5             \\ 
Finetune & FE-TResNet & ChineseFoodNet & 224   & 100    & 48         & Adam      & 1.E-04 & Cosine   & 1.E-5        & 0             \\ 
Finetune & FE-TResNet & CNFOOD-241     & 224   & 100    & 48         & Adam      & 1.E-04 & Cosine   & 1.E-5        & 0             \\ \hline
\end{tabular}
\begin{tablenotes} 
        \footnotesize             
        \item{Note: In the table mentioned,"Task" represents the type of problem the model is designed to solve. "Model" represents the different models used. "Datasets" represents the various datasets employed. "Input" refers to the size of the data fed into the model. "Epochs" indicates the number of training cycles. "Batch Size" refers to the number of samples fed into the model during each training iteration. "Optimizer" is used to update the model's weights. "LR" is learning Rate. "LR Decay" reduces the learning rate as training progresses. "Weight Decay" is regularization technique. "Warmup Epochs" stabilizes the training process and prevents large weight updates at the start.}         
\end{tablenotes}
\end{threeparttable}
}
\end{table*}

\subsection{Comparative Parameters}
In the realm of deep learning, assessing a model's excellence transcends the Conventional metrics such as parameter count, floating-point operations, and memory footprint. It also encompasses a suite of specific metrics that delineate the model's predictive prowess. Among these are Precision, which quantifies the ratio of true positive predictions to all positive predictions made by the model; Recall, which measures the proportion of actual positive cases correctly identified; and the F1 score, a harmonic mean that balances both Precision and Recall, offering a singular measure of a model's accuracy in binary classification contexts. For classification tasks, the evaluative criteria expand to include Top-1 Accuracy, reflecting the model's ability to precisely predict the most likely category, and Top-5 Accuracy, indicating the model's capacity to list the correct category within its top five predictions. The precise computational formulas for these metrics are articulated as follows:
\begin{equation}
    Precision=\frac{TP}{TP+FP} 
\end{equation}
\begin{equation}
    Recall=\frac{TP}{TP+FN} 
\end{equation}
\begin{equation}
   F1=2 \frac{Precision \times Recall}{Precision+Recall}
\end{equation}
\begin{equation}
    Top-1\ Accuracy=\frac{P1}{PA}
\end{equation}
\begin{equation}
    Top-5\ Accuracy=\frac{P5}{PA}
\end{equation}
In the context of deep learning, True Positives (TP), True Negatives (TN), False Positives (FP), and False Negatives (FN) are pivotal terms that define the outcomes of classification models.
In the context of the formulas, P1 denotes the count of instances where the model's most probable prediction aligns perfectly with the actual class. P5 refers to the number of cases where the true class is included within the model's top five predicted categories. Lastly, PA signifies the overall count of all instances evaluated in the classification task.
With these metrics and their corresponding formulas, one can discern the performance of various models in image classification tasks, particularly in fine-grained categorization. By comparing the experimental results across multiple models, it becomes evident which network architecture excels in fine-grained image classification tasks. The model demonstrating the most outstanding performance is then selected as the benchmark. Subsequently, the methodological enhancements are integrated into the training process, building upon the strengths of this benchmark model. This approach ensures a systematic and empirical method for identifying and refining the most effective deep learning architectures for the nuanced challenges of fine-grained classification. By leveraging the best-performing model as a foundation and incorporating the innovative techniques, this study aim to push the boundaries of what is achievable in the domain of image classification.
\subsection{Comparative experiments and analysis}
This paper have harnessed a suite of esteemed deep learning model architectures to assess two extensive datasets pertinent to Chinese cuisine. (1) The Residual Networks, ResNet50 \citep{r12} and ResNet101 \citep{r13}, are pivotal in mitigating the degradation phenomena within deep networks, averting issues of gradient vanishing and explosion throughout the training regimen; (2) The Dense Convolutional Networks, namely DenseNet121, DenseNet161, DenseNet169, and DenseNet201 \citep{r33}, amalgamate features across layers on the channel dimension to rival the efficacy of residual networks; (3) The TResNet architecture, comprising TResNet-M, TResNet-L, and TResNet-XL \citep{r1}, emerged from the quest to enhance the GPU training efficiency of ResNet models; (4) The EfficientNets, including EfficientNetB0 through B3 \citep{r36}, and its variant EfficientNetV2-S \citep{r44}, leverage compound scaling and AutoML to dynamically optimize CNNs across various dimensions; (5) The progression of Inception models, InceptionV2 \citep{r34}, InceptionV3, and InceptionV4 \citep{r34,r35}, has been geared towards the decoupling of Convolutions, harnessing parallel group Convolutions of assorted sizes to garner richer feature sets; (6) Xception \citep{r39,r45}, an innovative deep Convolutional architecture inspired by Inception, supplants traditional Convolutions with depthwise separable Convolutions; (7) MobileNet \citep{r37,r38}, designed for mobile applications, also relies on depthwise separable Convolutions, prioritizing model lightweightness in contrast to Xception's emphasis on performance.

To ensure a fair comparative analysis, uniform image preprocessing was applied to all datasets before experimentation, with consistent configurations maintained throughout. The model, FE-TResNet, was compared with the aforementioned renowned deep learning models on the ChineseFoodNet and CNFOOD-241 datasets. The experimental outcomes, detailed in Table \ref{tab:two}, include Parameters (Par.), Floating-Point operations (FL.), Memory Usage (Mem.), Top-1 Accuracy (Top-1 Acc.), and Top-5 Accuracy (Top-5 Acc.). The Top-1 Acc. results show that FE-TResNet outperforms its architectural counterparts in classification precision, adeptly capturing both global and local features and effectively discerning intra-class variations and inter-class similarities within food imagery. However, FE-TResNet also faces the inherent challenges of high parameter volume, substantial computational demand, and memory intensity, which may elongate training durations.

\begin{table*}[!ht]
\scriptsize
\centering
\caption{Each model is presented with 50 rounds of best results}
\label{tab:two}

\resizebox{\textwidth}{52mm}{
\begin{threeparttable} 
\begin{tabular}{llllllllllllll}
\hline
Methods & Resolution & & \multicolumn{5}{l}{ChineseFoodNet}                                                    & & \multicolumn{5}{l}{CNFOOD-241}  \\\cline{4-8} \cline{10-14}
        &           & & Par.(M) & FL.(G) & Mem.(MiB) & Top-1 Acc.(\%) & Top-5 Acc.(\%)      & & Par.(M) & FL.(G) & Mem.(MiB) & Top-1 Acc.(\%) & Top-5 Acc.(\%) \\ \hline
Resnet50                 & 224 $\times$ 224 && 23.93 & 4.12  &3870 & 78.24 & 95.65 &&23.93 & 4.12 & 3866 & 76.49  &94.70     \\
Resnet101  & 224 $\times$ 224 && 42.92 & 7.84  &5120               
& 78.76 & 95.81&& 42.92 & 7.84  & 5122  &76.78  & 95.02      \\
DenseNet121  & 224 $\times$ 224 &&7.17 & 2.88   & 5156  & 78.46 & 95.73&&7.17 & 2.88& 5154  & 76.51  & 94.70    \\
DenseNet161   & 224 $\times$ 224 &&26.93 &7.82 & 8348         & 79.78  & 96.24&&26.93 &7.82& 8366 & 78.42  & 95.61 \\
DenseNet169  & 224 $\times$ 224 &&12.83 &3.42 & 6002         &78.73  & 95.80  &&12.83 &3.42 & 5796 & 77.02 &95.11 \\
DenseNet201              & 224 $\times$ 224 &&18.49 &4.37 &7182        & 79.44 &96.05 &&18.49 &4.37 &7190 & 77.57 &95.38 \\
TResNet-M    & 224 $\times$ 224 && 29.76  &5.74   &2996      & 80.82   & 96.41  &&29.76  &5.74    & 2996 & 78.48 &95.76 \\
TResNet-L    & 224 $\times$ 224 && 54.06 &10.88 &4366   & 80.71   & 96.50     && 54.06 &10.88 & 4362 & 79.31& 95.90 \\
TResNet-XL    & 224 $\times$ 224&& 76.33&15.17 &5512           &80.85       & 96.72&& 76.33&15.17 & 5524 & 79.85 & 96.12  \\
EfficientNetB0 & 224 $\times$ 224 &&4.27 &0.40 &3654 & 78.69  &95.87  &&4.27 &0.40  &3660 &76.33&95.01   \\
EfficientNetB1    & 240 $\times$ 240  &&6.78 &0.59 & 4550             & 78.39    &95.73   &&6.78 &0.59 & 4548 & 76.49 & 95.00   \\
EfficientNetB2 & 260 $\times$ 260  &&7.99 &0.68   &4690     & 78.79   &95.84 &&7.99 &0.68 &4690 & 77.20 & 95.37 \\
EfficientNetB3 & 300 $\times$ 300  &&11.02 &0.99 & 5676  & 79.16    &95.99    &&11.02 &0.99  & 5676&77.58 & 95.47\\
EfficientNetv2-S & 224 $\times$ 224  &&20.44     &2.87     &5494       & 81.11      &96.69   &&20.44     &2.87 &5500 &79.61 & 96.22\\
InceptionV2       & 299 $\times$ 299 &&54.63 &6.5 & 4786   & 78.19      &95.55    &&54.63 &6.5 &4786&76.86 &95.07 \\
InceptionV3 & 299 $\times$ 299 &&22.21 &2.85 & 3058      & 75.87      & 94.61    &&22.21 &2.85 &3056 & 74.28 & 93.80 \\
InceptionV4 & 299 $\times$ 299 &&41.46 &12.31 &4370    & 77.54        & 95.26    &&41.46 &12.31 & 4334 &76.30 & 94.78 \\
MoblienetV2              & 224 $\times$ 224 &&2.49 &0.32 &3312        &74.38    &94.26  &&2.49 &0.32 & 3310 & 71.31 &93.37 \\
MoblienetV3-s         & 224 $\times$ 224 &&1.73 &0.01 &1720      &71.19    &92.60 &&1.73 &0.01  &1718& 67.96       & 91.03     \\
MoblienetV3-l         & 224 $\times$ 224 &&4.47 &0.23 &2592      &76.44    & 94.98   &&4.47 &0.23 &2592& 74.30  & 93.90        \\
Xception41               & 299 $\times$ 299 &&25.37 &5.03 &6460  &78.41    &95.55  &&25.37 &5.03  & 6104 &76.60 & 95.02\\
Xception65              & 299 $\times$ 299 &&38.30 &7.57 &7876  &78.37    &95.84     &&38.30 &7.57 & 7952 & 77.09 & 95.06\\
Xception71              & 320 $\times$ 320 &&40.72  &9.84 &9852    & 79.06& 95.96 &&40.72  &9.84& 9769 & 77.20  &95.18\\
FE-TResNet(study)  & 224 $\times$ 224 && 82.95&15.82 &8006           &81.37       & 97.86&& 82.95&15.82 & 8006  &80.29  & 97.97\\\hline
\end{tabular}

\begin{tablenotes} 
        \footnotesize             
        \item{Note: In the table, "Par.(M)" denotes the number of parameters in millions (Params). "FL(G)" represents the number of floating-point operations required for the model to execute, in billions (Floating Point Operations). "Mem.(MiB)" refers to the memory usage of the model during operation, in megabytes (MegaBytes). "Top-1 Acc.(\%)" and "Top-5 Acc.(\%)" indicate the accuracy of the model's prediction matching the actual category and the accuracy of the actual category being included in the top five most likely categories predicted by the model, respectively.}         
\end{tablenotes}
\end{threeparttable}
}
\end{table*}

\subsection{Ablation experiment and analysis}
The results in Table \ref{tab:two} for Top-1 Acc. and Top-5 Acc. clearly show that, among the ChineseFoodNet and CNFOOD-241 datasets, the TResNet-XL model outperforms others in fine-grained image classification, a testament to its refined architecture derived from ResNet50 and the bolstering presence of the SE module. Its memory consumption is relatively low among the seven models, albeit with a higher volume of parameters. As a whole, TResNet demonstrates pronounced superiority in the realm of fine-grained image classification tasks. Enhancing the TResNet-XL model's capabilities, this study have integrated the StyleRM module, which accentuates stylistic features, and the DCA module, facilitating inter-pixel connectivity across the model's channels. The Top-1  Acc. results in Table \ref{tab:three} show that the incorporation of these modules has markedly improved the TResNet-XL model's fine-grained classification capabilities on the datasets, substantiating the precision of the method in the realm of food image classification. Upon examining the Parameters (Par.), Floating-Point operations (FL.), and Memory usage (Mem.), the method does not impose a significant computational burden or parameter increment on the foundational model, thereby underscoring the formidable adaptability of the FE-TResNet model.

In pursuit of a comprehensive evaluation of the efficiency of the proposed methodology, this study expanded the evaluation to encompass the F1 score, Precision, and Recall on the ChineseFoodNet and CNFOOD-241 datasets for ResNet101, DenseNet161, EfficientNetB2, EfficientNetB3, InceptionV3, Xception65, and the own FE-TResNet method. The results presented in Table \ref{tab:four} indicate that the method surpasses the other seven models across these metrics, indicating an adept balance between the identification of positive samples and the exclusion of negative ones, and an effective utilization of data features for distinguishing between categories. In summary, the FE-TResNet model not only excels in precision and adaptability but also demonstrates robust generalization capabilities, adeptly discerning minute variations within sample data. This establishes the FE-TResNet model as a potent solution for the fine-grained classification of food images.

\begin{table*}[!ht]
\scriptsize
\centering
\caption{Ablation experiment of FE-TResNet on the ChineseFoodNet and CNFOOD-241 datasets.}
\label{tab:three}
\setlength{\tabcolsep}{2mm}{
\begin{threeparttable} 
\begin{tabular}{lllllllllllllll}
\hline
\multicolumn{2}{l}{Methods} &  & \multicolumn{5}{l}{ChineseFoodNet}                      &  & \multicolumn{5}{l}{CNFOOD-241}      \\ \cline{1-2} \cline{4-8} \cline{10-14}
StyleRM            & DCA               & & Par.(M) & FL.(G) & Mem.(MiB) & Top-1 Acc.(\%) & Top-5 Acc.(\%)      & & Par.(M) & FL.(G) & Mem.(MiB) & Top-1 Acc.(\%) & Top-5 Acc.(\%) \\ \hline
\XSolidBrush        & \XSolidBrush  && 76.33&15.17 &5512           &80.85       & 96.72&& 76.33&15.17 & 5524 & 79.85 & 96.12  \\
\CheckmarkBold        & \XSolidBrush  && 76.33&15.17 &6192      &81.22       & 97.13&& 76.33&15.17 &6178   &80.15  & 97.23        \\
\XSolidBrush          & \CheckmarkBold  &&82.94&15.82 &7362     &80.91 &97.03&&82.94&15.82 &7360  &79.92& 96.56      \\
\CheckmarkBold        & \CheckmarkBold  && 82.95&15.82 &8006           &81.37       & 97.86&& 82.95&15.82 & 8006  &80.29  & 97.97      \\ \hline              
\end{tabular}
\begin{tablenotes} 
        \footnotesize             
        \item{Note: In the table, "Par.(M)" denotes the number of parameters in millions (Params). "FL(G)" represents the number of floating-point operations required for the model to execute, in billions (Floating Point Operations). "Mem.(MiB)" refers to the memory usage of the model during operation, in megabytes (MegaBytes). "Top-1 Acc.(\%)" and "Top-5 Acc.(\%)" indicate the accuracy of the model's prediction matching the actual category and the accuracy of the actual category being included in the top five most likely categories predicted by the model, respectively.}         
\end{tablenotes}
\end{threeparttable}
}
\end{table*}

\begin{table*}[!ht]
\scriptsize
\centering
\caption{Analysis of other aspects of image classification performance.}
\label{tab:four}
\setlength{\tabcolsep}{2mm}{
\begin{threeparttable} 
\begin{tabular}{lccccccccccc}
\cline{1-12} 
Methods        & Resolution &  & \multicolumn{4}{l}{ChineseFoodNet}                   &  & \multicolumn{4}{l}{CNFOOD-241}                               \\ \cline{4-7} \cline{9-12} 
               &            &  & Accuracy(\%) & F1(\%) & Precision(\%) & Recall(\%)             &  & Accuracy(\%) & F1(\%) & Precision(\%) & Recall(\%) \\ \hline
Resnet101       & 224 $\times$ 224  &&78.76  & 76.34&77.25&76.13 && 76.78 &75.24         & 75.34 &76.24 \\
DenseNet161    & 224 $\times$ 224 &&79.78  & 77.60& 78.43  & 77.26                     &&78.42& 76.88 &76.81 &77.81 \\
EfficientNetB2 & 260 $\times$ 260 &&78.79 &76.69 &77.62&76.48  &&77.20&75.55&75.58 & 76.73 \\
EfficientNetB3     & 300 $\times$ 300  &&79.16 & 77.01  & 77.67 & 77.02 && 77.58         & 76.08 &76.19 & 77.11 \\
InceptionV3   & 299 $\times$ 299  && 75.87  &73.48&74.41 &73.16&& 74.28&73.59 & 74.54    &73.21 \\
Xception65    & 299 $\times$ 299  &&78.37 &76.80 &77.51  &76.48   && 77.09 &75.99  &76.18     & 76.91 \\
FE-TResNet(study)  & 224 $\times$ 224  &&81.37  &80.37 &80.88 &80.18 && 80.29 & 79.61  &79.51      & 80.27 \\ \hline
\end{tabular}
\begin{tablenotes} 
        \footnotesize             
        \item{Note: In the table, "Accuracy(\%)" is consistent with the aforementioned "Top-1 Acc.(\%)". "Precision(\%)" represents the precision of the model's predictions that correctly identify the positive class. "Recall(\%)" indicates the recall rate, which is the proportion of the actual positive class that is correctly predicted by the model. "F1(\%)" denotes the harmonic mean of precision and recall.}         
\end{tablenotes}
\end{threeparttable}
}
\end{table*}

\section{Conclusion and future work}
\label{}
This study introduces a high-precision food image classification method based on TResNet, named FE-TResNet. Through feature enhancement technology, the method can accurately capture subtle features in fine-grained food images, effectively handling food images that are similar in shape but differ in subtle details. To enhance the model's generalization ability, automated data augmentation techniques were utilized. Additionally, incorporating the StyleRM module significantly improved the model's performance in extracting low-level features by assigning different weights to the data. Meanwhile, leveraging DCA technology integrated multi-channel information, which not only promoted the model's capability in multi-scale feature fusion but also addressed the need for processing a large number of pixel points in high-resolution images, thereby enhancing the model's feature extraction performance at the high-level stage. Experimental results on complex food datasets ChineseFoodNet and CNFOOD-241 showed superior accuracy for the model in fine-grained image classification tasks compared to other self-supervised models. Results indicated that the model could better cope with intra-class variability and inter-class similarity, achieving more precise category prediction. Therefore, this method has the potential to help people record and adjust dietary habits, and improve health levels, by identifying and classifying food images. Although FE-TResNet has shown excellent classification performance on food datasets, it remains untested on image datasets in other domains. In the future, based on the foundation of food classification, this paper aim to enhance the model to be more extensible and apply it to a broader range of computer vision fields to solve more complex classification problems.

% \section*{Declaration of Competing Interest}
% The authors declare that they have no known competing financial interests or personal relationships that could have appeared to influence the work reported in this paper.
\section*{Ethical approval}
The study does not involve any human or animal testing.
\section*{Disclosure statement}
No potential conflict of interest was reported by the author(s).
\section*{Data availability}
The data that support the findings of this study are available from the corresponding author, upon reasonable request.

%Bibliography
\bibliographystyle{unsrt}  
\bibliography{templateArxiv}

\nocite{*}

\end{document}